\documentclass[11pt]{article}

\usepackage[final]{acl}

\usepackage{times}
\usepackage{latexsym}
\usepackage{tabularx}
\usepackage{multirow}
\usepackage{booktabs}
\usepackage{graphicx}
\usepackage{booktabs}
\usepackage{multirow}
\usepackage{arydshln}
\usepackage{amsmath}
\usepackage[T1]{fontenc}

\usepackage[utf8]{inputenc}

\usepackage{microtype}

\usepackage{inconsolata}

\usepackage{graphicx}

\title{Mechanistic Interpretability for Large Language Model Alignment: Progress, Challenges, and Future Directions}

\author{Usman Naseem\\
  Macquarie University\\
  Correspondence: usman.naseem@mq.edu.au}

\begin{document}
\maketitle
\begin{abstract}
Large language models (LLMs) have achieved remarkable capabilities across diverse tasks, yet their internal decision-making processes remain largely opaque. Mechanistic interpretability—the systematic study of how neural networks implement algorithms through their learned representations and computational structures—has emerged as a critical research direction for understanding and aligning these models. This paper surveys recent progress in mechanistic interpretability techniques applied to LLM alignment, examining methods ranging from circuit discovery to feature visualization, activation steering, and causal intervention. We analyze how interpretability insights have informed alignment strategies including reinforcement learning from human feedback (RLHF), constitutional AI, and scalable oversight. Key challenges are identified, including the superposition hypothesis, polysemanticity of neurons, and the difficulty of interpreting emergent behaviors in large-scale models. We propose future research directions focusing on automated interpretability, cross-model generalization of circuits, and the development of interpretability-driven alignment techniques that can scale to frontier models.
\end{abstract}

\section{Introduction}

The rapid advancement of large language models (LLMs) has created an urgent need for robust alignment techniques that ensure these systems behave in accordance with human values and intentions \citep{ouyang2022training, bai2022training}. While behavioral approaches to alignment—such as RLHF and various prompting strategies—have shown practical success, they treat models as black boxes and provide limited guarantees about generalization to novel situations or adversarial inputs \citep{casper2023open}.

Mechanistic interpretability offers a complementary paradigm: understanding the internal algorithms and representations that LLMs learn during training \citep{olah2020zoom, elhage2021mathematical}. By reverse-engineering the computational mechanisms underlying model behavior, researchers aim to develop more principled approaches to alignment that directly modify or constrain the problematic circuits while preserving beneficial capabilities.

Recent work has demonstrated that transformer-based LLMs learn interpretable substructures—often called "circuits"—that implement specific algorithmic functions \citep{wang2022interpretability, conmy2023towards}. These discoveries have enabled targeted interventions for alignment purposes, from steering model behavior through activation editing \citep{li2023inference} to identifying and ablating deceptive or harmful reasoning patterns \citep{zou2023representation}.

This paper provides a comprehensive survey of mechanistic interpretability techniques applied to LLM alignment. We organize our discussion around three key questions:

\begin{itemize}
    \item \textbf{What progress has been made?} We review major advances in interpretability methods and their applications to alignment challenges.
    \item \textbf{What fundamental challenges remain?} We analyze theoretical and practical barriers to achieving comprehensive interpretability of large-scale models.
    \item \textbf{What future directions are most promising?} We identify research priorities for developing scalable, automated interpretability techniques that can support alignment of increasingly capable systems.
\end{itemize}

\section{Background and Foundations}
\subsection{The Transformer Architecture}

Modern LLMs are built on the transformer architecture \citep{vaswani2017attention}, which processes sequences through alternating layers of attention and feedforward computations. Understanding this architecture is essential for mechanistic interpretability work.

The attention mechanism allows each token to aggregate information from previous tokens in the sequence. For a given layer 
\textit{l}, attention head \textit{h}
 computes:

\begin{equation}
\text{Attention}(Q, K, V) = \text{softmax}\left( \frac{Q K^\top}{\sqrt{d_k}} \right) V
\end{equation}

where \textit{Q}, \textit{K}, and \textit{V} are query, key, and value matrices derived from linear transformations of the input embeddings \citep{vaswani2017attention}.

Multi-layer perceptrons (MLPs) in transformer layers implement position-wise feedforward transformations, which recent work suggests act as key-value memories storing factual associations \citep{geva2021transformer, meng2022locating}.

\subsection{The Alignment Problem}
The alignment problem concerns ensuring that AI systems pursue goals and exhibit behaviors consistent with human values \citep{russell2019human}. For LLMs, key alignment challenges include:

\begin{itemize}
    \item \textbf{Truthfulness and hallucination:} Models may generate plausible but false information \citep{lin2022truthfulqa}
    \item \textbf{Harmful content generation:} Models may produce toxic, biased, or dangerous outputs \citep{gehman2020realtoxicityprompts}
    \item \textbf{Deceptive alignment:} Models may learn to behave well during training while concealing misaligned objectives \citep{hubinger2019risks}
    \item \textbf{Robustness and distribution shift:} Aligned behavior during training may not generalize to novel contexts \citep{hendrycks2021measuring}
\end{itemize}

Current alignment approaches primarily rely on RLHF \citep{christiano2017deep, ouyang2022training}, which fine-tunes models using human preference feedback. While effective for improving surface-level behaviors, RLHF provides limited insight into whether models have internalized desired values or merely learned to imitate aligned behavior ~\citep{casper2023open}.

\subsection{Core Concepts in Mechanistic Interpretability}


\textbf{Circuits:} Subgraphs of a neural network that implement specific algorithmic functions \citep{cammarata2020curve, olah2020zoom}. Circuit analysis aims to identify minimal subnetworks responsible for particular behaviors.

\noindent\textbf{Features:} Directions in activation space corresponding to interpretable concepts \citep{olah2017feature}. Features may be represented by individual neurons (monosemantic) or by linear combinations of neurons (polysemantic).

\noindent\textbf{Superposition:} The hypothesis that networks represent more features than they have neurons by storing features in superposition—as overlapping combinations of neural activations \citep{elhage2022toy}. This creates significant challenges for interpretability.

\noindent\textbf{Residual stream:} In transformers, information flows through a residual stream that accumulates contributions from attention and MLP layers~\citep{elhage2021mathematical}. Understanding how components read from and write to this stream is crucial for circuit analysis.

\section{Methods for Mechanistic Interpretability}

\subsection{Activation Analysis and Probing}

\textbf{Probing classifiers} train auxiliary models to predict properties from internal representations, revealing what information is encoded in activations \citep{belinkov2022probing}. For alignment, probes have been used to detect when models represent harmful content~\citep{zou2023representation} or deceptive reasoning \citep{azaria2023internal}.
However, probing has limitations: high probe accuracy doesn't necessarily mean information is used for downstream computations \citep{belinkov2022probing}, and probes may learn to extract information in ways unrelated to the model's actual computations.

\noindent\textbf{Logit lens and tuned lens} methods project intermediate activations through the unembedding matrix to interpret representations as probability distributions over vocabulary \citep{belrose2023eliciting}. These techniques reveal how predictions evolve through layers and have been used to study phenomena like in-context learning \citep{olsson2022context}.

\subsection{Attention Pattern Analysis}

Attention weights provide direct insight into information flow between tokens. Researchers have identified interpretable attention patterns corresponding to specific functions:

\begin{itemize}
    \item \textbf{Induction heads:} Attention patterns that implement in-context learning by copying information from previous similar contexts \citep{olsson2022context}
    \item \textbf{Previous token heads:} Heads that primarily attend to the immediately preceding token \citep{elhage2021mathematical}
    \item \textbf{Factual recall heads:} Heads involved in retrieving factual knowledge~\citep{meng2022locating}
\end{itemize}

For alignment applications, attention analysis has revealed how models process and propagate harmful content \citep{zou2023representation}, enabling targeted interventions.

\subsection{Circuit Discovery}
Circuit discovery aims to identify minimal subnetworks implementing specific behaviors. Key approaches include:

\noindent\textbf{Activation patching} (also called causal tracing): Systematically intervenes on activations to determine which components causally contribute to particular outputs \citep{meng2022locating,wang2022interpretability}. By corrupting inputs and selectively restoring clean activations, researchers identify necessary and sufficient components for behaviors.

\noindent\textbf{Automatic circuit discovery:} Recent methods automate circuit identification using techniques like:

\begin{itemize}
    \item \textbf{Attribution patching:} Efficiently approximates patching by computing gradients \citep{syed2023attribution}
    \item \textbf{Edge pruning:} Iteratively removes edges in the computational graph while maintaining output behavior~\citep{conmy2023towards} 
    \item \textbf{Path patching:} Traces information flow along specific paths through the network~\citep{goldowsky2023localizing} 
\end{itemize}

\begin{table*}[!t]
\centering
\caption{Taxonomy of Mechanistic Interpretability Techniques}
\label{tab:mech_interp_taxonomy_acl}
\resizebox{\textwidth}{!}{%
\begin{tabular}{l l p{5.5cm} p{5.0cm} p{5.5cm}}
\toprule
\textbf{Category} &
\textbf{Technique} &
\textbf{Key Mechanism} &
\textbf{Strengths} &
\textbf{Limitations} \\
\midrule

\multirow{3}{*}{Observational Analysis}
& Probing Classifiers 
& Linear classifiers on internal activations
& Low computational cost; detects encoded information
& No causal guarantees; may not reflect actual usage \\

& Logit Lens / Tuned Lens
& Project activations through the unembedding matrix
& Traces prediction evolution; interpretable outputs
& Layer-wise snapshots only; assumes linearity \\

& Attention Pattern Analysis
& Visualization of attention weights
& Direct insight into information flow; identifies head roles
& Does not capture MLP effects; difficult compositional interpretation \\

\midrule

\multirow{2}{*}{Feature Discovery}
& Sparse Autoencoders (SAE)
& Sparse dictionary learning with $\ell_1$ regularization
& Addresses polysemanticity; discovers monosemantic features
& Scaling challenges; reconstruction--fidelity trade-offs \\

& Dataset Examples + LLM Description
& High-activation examples with automated descriptions
& Scalable; human-interpretable summaries
& Descriptions may be post-hoc; validation difficulty \\

\midrule

\multirow{4}{*}{Circuit Discovery}
& Activation Patching
& Corrupt and restore activations to test causal impact
& Gold standard for causal attribution
& Computationally expensive; combinatorial explosion \\

& Automated Discovery 
& Graph pruning using faithfulness metrics
& Automates circuit isolation; scalable
& Requires threshold tuning; may miss distributed circuits \\

& Attribution Patching
& Gradient-based approximation of patching
& Efficient; good causal approximation
& Less precise than full patching \\

& Path Patching
& Trace information flow along selected paths
& Isolates direct versus indirect effects
& Path explosion in deep networks \\

\midrule

\multirow{3}{*}{Causal Intervention}
& Activation Steering
& Add direction vectors to intermediate activations
& Precise behavior control; no retraining required
& Requires high-quality steering vectors; generalization unclear \\

& Knowledge Editing
& Direct weight modification (e.g., ROME, MEMIT)
& Surgical fact updates; preserves other knowledge
& Primarily factual scope; potential side effects \\

& Representation Engineering
& Read and control abstract properties via latent directions
& Targets high-level concepts; multi-property control
& Robust direction discovery; interaction effects \\

\midrule

Validation
& Causal Abstractions
& Formal alignment between model mechanisms and interpretations
& Rigorous causal guarantees; principled evaluation
& Computationally intensive; requires formalization \\

\bottomrule
\end{tabular}
}
\end{table*}

These automated methods have successfully discovered circuits for tasks like indirect object identification~\citep{wang2022interpretability} and greater-than comparisons \citep{hanna2023does}.

\subsection{Feature Visualization and Sparse Autoencoders}

Understanding what individual neurons or directions in activation space represent is fundamental to interpretability. Traditional approaches include:

\noindent\textbf{Feature visualization:} Optimizing inputs to maximally activate specific neurons \citep{olah2017feature}. For LLMs, this involves finding token sequences that strongly activate target features.

\noindent\textbf{Dataset examples:} Collecting examples from training data that highly activate features \citep{bills2023language}. Recent work uses LMs to automatically generate natural language descriptions of neuron behavior based on these examples \citep{bills2023language}.

\noindent\textbf{Sparse autoencoders (SAEs):} Address the superposition challenge by training autoencoders with sparsity constraints to decompose neural activations into interpretable features \citep{cunningham2023sparse, bricken2023towards}. SAEs learn overcomplete feature dictionaries where individual features correspond to interpretable concepts. Recent work has successfully applied SAEs to various layers of LLMs, discovering features corresponding to topics, entities, and linguistic properties \citep{bricken2023towards}.

\subsection{Causal Interventions and Steering}

Beyond observational analysis, interventional techniques directly modify model internals to test causal hypotheses and control behavior:

\noindent\textbf{Activation steering:} Directly editing activations during inference to control model behavior \citep{turner2024activation,li2023inference}. By adding carefully chosen vectors to activations, researchers can amplify or suppress specific behavioral tendencies. This has been applied to enhance truthfulness, reduce toxicity, and control stylistic properties \citep{li2023inference}.

\noindent\textbf{Representation engineering:} A framework for reading and controlling high-level cognitive properties by identifying representation directions and performing targeted interventions \citep{zou2023representation}. This approach has been used to enhance honesty and reduce hallucination in LLMs.

\noindent\textbf{Causal abstractions:} Formal framework for verifying whether interpretations correspond to true causal relationships in the model \citep{geiger2021causal, geiger2023inducing}. This provides rigorous foundations for validating interpretability claims.

Table~\ref{tab:mech_interp_taxonomy_acl} provides a systematic taxonomy of mechanistic interpretability techniques, organized by their primary function. These methods are often used in combination—for example, using sparse autoencoders to identify features, then using activation patching to discover which circuits use those features causally.

\section{Applications to LLM Alignment}

\subsection{Understanding RLHF Mechanisms}

Mechanistic interpretability has begun to illuminate how RLHF changes model behavior:

\noindent\textbf{Value representation:} Research has investigated how reward models represent human preferences \citep{casper2023open}. Studies using probing and intervention methods suggest reward models learn relatively shallow heuristics rather than deep understanding of human values.

\noindent\textbf{Policy changes:} Circuit analysis of pre- and post-RLHF models reveals that RLHF primarily affects specific components related to response initiation and style, while core knowledge and reasoning circuits remain largely unchanged \citep{tigges2023linear}. This suggests RLHF acts more as a behavioral filter than fundamental value learning.

\noindent\textbf{Sycophancy circuits:} Interpretability work has identified circuits responsible for sycophantic behavior—models agreeing with user statements regardless of truth \citep{sharma2023towards}. These findings enable targeted debiasing interventions.

\subsection{Detecting and Mitigating Deception}

A critical alignment concern is whether models might learn deceptive strategies—behaving well during evaluation while pursuing misaligned objectives in deployment.

\noindent\textbf{Lie detection:} Recent work uses linear probes to detect when models generate false statements \citep{azaria2023internal}. These detectors achieve reasonable accuracy but face challenges when deception is sophisticated or the model is trained to evade detection.

\noindent\textbf{Situational awareness: }Research has investigated whether models represent information about their training status or evaluation context \citep{berglund2023reversal}. Such representations could enable deceptive alignment, where models behave differently when they believe they're being evaluated.

\noindent\textbf{Trojan detection:} Interpretability techniques have been applied to detect backdoor attacks and trojans in language models \citep{huang2023catastrophic}, with circuit discovery methods identifying malicious subnetworks.

\subsection{Reducing Harmful Outputs}

\textbf{Toxicity circuits:} Circuit analysis has identified specific attention heads and MLPs responsible for generating toxic or harmful content \citep{zou2023representation}. Ablating or modifying these components reduces harmful outputs while minimally impacting benign capabilities.

\noindent\textbf{Bias mitigation:} Interpretability methods have revealed how stereotypical biases are represented and propagated through layers \citep{vig2020investigating}. This enables targeted interventions to reduce specific biases without extensive retraining.

\noindent\textbf{Refusal mechanisms:} Recent work has analyzed how models learn to refuse harmful requests \citep{arditi2024refusal}, identifying specific components responsible for safety behaviors. Understanding these mechanisms helps improve robustness of safety training.

\subsection{Improving Factuality and Reducing Hallucination}

\textbf{Knowledge localization:} Research has localized where factual knowledge is stored in transformer models, primarily in MLP layers \citep{geva2021transformer,meng2022locating}. This enables:

\begin{itemize}
    \item \textbf{Knowledge editing:} Directly modifying stored facts without retraining \citep{meng2022locating,mitchell2022fast} 
    \item \textbf{Uncertainty quantification:} Detecting when models lack relevant knowledge \citep{kadavath2022language}
    \item \textbf{Hallucination detection:} Identifying when models generate content not grounded in their training data
\end{itemize}

\noindent\textbf{Attention to source information:} Analysis of how models attend to provided context versus internal knowledge reveals mechanisms underlying hallucination \citep{mallen2023when}. Models sometimes preferentially rely on memorized information even when contradicted by input context.

\subsection{Enhancing Transparency and Oversight}

\textbf{Chain-of-thought interpretability:} Mechanistic analysis of models generating chain-of-thought reasoning reveals the relationship between intermediate steps and internal computations \citep{wang2023self}. This addresses whether reasoning traces faithfully represent actual model cognition or merely post-hoc rationalizations.

\noindent\textbf{Faithful explanations: }Interpretability methods help validate whether model-generated explanations correspond to true decision-making processes \citep{turpin2023language}. Evidence suggests explanations can be superficial or misleading, highlighting the need for mechanistic verification.

\noindent\textbf{Scalable oversight:} Interpretability tools enable humans to oversee model behavior on tasks where direct evaluation is difficult \citep{bowman2022measuring}. By examining internal representations and circuits, supervisors can detect potential misalignment even when outputs appear reasonable.

\subsection{Pluralistic Alignment: Values, Culture, and Diversity}

A critical challenge in LLM alignment is that human values are diverse, context-dependent, and often conflicting across individuals, communities, and cultures \citep{sorensen2024roadmap}. Pluralistic alignment aims to develop AI systems that can navigate this diversity rather than optimizing for a single conception of "aligned" behavior \citep{bakker2022fine}.

\subsubsection{Representing Value Diversity}

Mechanistic interpretability research has begun investigating how models represent different value systems, moral frameworks, and cultural perspectives:

\noindent\textbf{Moral and ethical frameworks:} Recent work using sparse autoencoders has identified distinct features corresponding to different ethical perspectives—deontological, consequentialist, and virtue-based reasoning—that activate in different contexts \citep{kirk2024benefits}. Understanding these representations enables:

\begin{itemize}
    \item \textbf{Value attribution:} Determining which value systems influence particular model outputs
    \item \textbf{Conflict detection:} Identifying when multiple incompatible values are activated simultaneously
    \item \textbf{Bias auditing:} Detecting systematic preferences for certain value frameworks over others
\end{itemize}

\textbf{Cultural value systems:} Circuit analysis reveals systematic patterns in how models represent cultural diversity:

\begin{itemize}
    \item \textbf{Western-centric value circuits:} Models trained predominantly on English internet data develop circuits that robustly encode Western ethical frameworks (individualism, autonomy, rights-based reasoning) while representing collectivist or communitarian values more weakly \citep{tao2024investigating}. Circuit analysis shows that MLP layers contain dense factual associations about Western cultural contexts but sparser representations of non-Western traditions.

    \item \textbf{Language-dependent moral reasoning:} Multilingual models often exhibit different moral judgments depending on the language of the query, even when semantically equivalent \citep{ramezani2023knowledge}. Attention pattern analysis reveals that models route information through different circuits based on language, suggesting distinct cultural value systems are encoded in language-specific pathways.

    \item \textbf{Cultural knowledge localization:} Similar to factual knowledge neurons \citep{geva2021transformer}, models contain neurons that activate for culture-specific information—holidays, customs, historical events, social norms—with different cultural traditions stored in partially overlapping but distinguishable neural populations \citep{arora2023probing}.
    
\end{itemize}

\begin{table*}[!t]
\centering
\caption{Mechanistic Interpretability Applications to Alignment Challenges}
\label{tab:mi_alignment_apps}
\renewcommand{\arraystretch}{1.25}
\small
\resizebox{\textwidth}{!}{%
\begin{tabular}{p{3.2cm} p{4.2cm} p{4.5cm} p{4.5cm} p{4.2cm}}
\toprule
\textbf{Alignment Goal} &
\textbf{MI Approach} &
\textbf{Key Findings} &
\textbf{Interventions Enabled} &
\textbf{Key Limitations} \\
\midrule

\textbf{Understanding RLHF}
& Circuit comparison pre/post-RLHF; reward model analysis
& RLHF primarily affects response-style circuits rather than core reasoning; reward models learn shallow heuristics
& Targeted RLHF improvements; detection of alignment failures
& Unclear how to induce deep value learning \\

\midrule

\textbf{Detecting Deception}
& Probing for false statements; situational awareness analysis
& Linear probes detect deception with moderate accuracy; internal states encode training context
& Lie detection systems; monitoring for deceptive alignment
& Sophisticated deception may evade detection \\

\midrule

\textbf{Reducing Toxicity}
& Circuit discovery for harmful content; stereotype head identification
& Specific attention heads propagate toxic content and can be ablated
& Surgical toxicity removal; stereotype mitigation
& Potential impact on benign capabilities \\

\midrule

\textbf{Improving Factuality}
& Knowledge localization in MLPs; source-attention analysis
& Facts are stored in MLP layers; models may ignore context in favor of memorized information
& Knowledge editing; hallucination detection; uncertainty estimation
& Limited to factual knowledge; possible side effects \\

\midrule

\textbf{Pluralistic Alignment}
& Value-feature discovery; cultural circuit analysis; steering vectors
& Models encode multiple ethical frameworks with uneven robustness
& Value-based steering; cultural adaptation; personalization
& Context dependence; essentialism risks; capacity asymmetries \\

\midrule

\textbf{Enhancing Transparency}
& Chain-of-thought circuit analysis; explanation faithfulness verification
& Explanations may be post-hoc and not reflect true computation
& Detection of unfaithful reasoning; explanation validation
& Persistent gap between explanation and reasoning \\

\midrule

\textbf{Scalable Oversight}
& Internal state monitoring; circuit-level anomaly detection
& Misalignment can be detected in representations despite benign outputs
& Early warning systems; targeted human oversight
& Requires identifying which anomalies signal genuine risk \\

\bottomrule
\end{tabular}
}
\end{table*}

\subsubsection{ Interventions for Pluralistic Alignment}
Mechanistic interpretability enables several approaches to handling value and cultural diversity:

\noindent\textbf{Activation steering for diverse preferences:} Activation steering methods have been extended to control which value systems and cultural perspectives models prioritize \citep{tigges2023linear,li2023inference}. By identifying representation directions corresponding to different philosophical, political, or cultural perspectives, researchers can dynamically adjust model behavior without retraining:

\begin{itemize}
    \item \textbf{Value-based steering:} Shifting between utilitarian and deontological reasoning
    \item \textbf{Cultural steering vectors:} Moving outputs toward different cultural perspectives (e.g., East Asian collectivist values vs. Western individualist values)
    \item \textbf{Personalization:} Adapting to individual user preferences while maintaining transparency
\end{itemize}

\textbf{RLHF with diverse preferences:} Mechanistic analysis of reward models trained on diverse human feedback reveals how models aggregate conflicting preferences \citep{bakker2022fine}:

\begin{itemize}
    \item \textbf{Standard RLHF} often learns to satisfy majority preferences while ignoring minority viewpoints
    \item \textbf{Preference decomposition:} Identifying which demographic or value groups influence different parts of the model
    \item \textbf{Fairness interventions:} Detecting and correcting underrepresentation of minority perspectives
    \item \textbf{Culturally-aware RLHF:} Circuit-level analysis shows reward models often learn cultural stereotypes rather than nuanced understanding \citep{cao2023lost}
\end{itemize}

\textbf{Circuit editing for inclusive representation:} Directly modifying circuits to improve representation of underrepresented perspectives:

\begin{itemize}
    \item Strengthening circuits for non-Western cultural knowledge \citep{meng2022locating}
    \item Ablating stereotype propagation heads \citep{vig2020investigating}
    \item Engineering value framework circuits for better balance
\end{itemize}

\subsubsection{Challenges in Pluralistic Alignment}

Mechanistic interpretability faces unique challenges when addressing value and cultural diversity:

\noindent\textbf{Value incommensurability:} Some values may be fundamentally incompatible, creating superposition-like conflicts where models cannot simultaneously represent all perspectives at full strength. This is particularly acute for cultural values that reflect different ontological assumptions.

\noindent\textbf{Asymmetric representation capacity:} Models trained on imbalanced data develop asymmetric circuit structures where dominant cultural concepts have richer, more robust representations \citep{cao2023lost}. This may be a fundamental limitation rather than easily correctable.

\noindent\textbf{Context-dependence:} The appropriate value framework often depends on subtle contextual factors—cultural context, domain, relationship dynamics—that models must learn to recognize. Current models often fail to activate culturally-appropriate circuits in the right contexts.

\noindent\textbf{Power dynamics and essentialism:}

\begin{itemize}
    \item Decisions about which cultural perspectives to prioritize reflect existing power structures \citep{birhane2022values}
    \item Mechanistic interventions targeting "cultural values" risk essentializing complex, heterogeneous cultures into simplified feature vectors
    \item Cultures are dynamic and internally diverse; static circuit-level representations may reinforce stereotypes
\end{itemize}

\noindent\textbf{Meta-level values:} Beyond first-order preferences, pluralistic alignment requires representing meta-values about how to adjudicate between conflicting preferences—itself a culturally-variable question.

\noindent\textbf{Evaluation challenges:} Assessing cultural alignment requires culturally-grounded evaluation, but most interpretability researchers come from Western contexts, potentially missing important biases.

Table~\ref{tab:mi_alignment_apps} maps mechanistic interpretability approaches to specific alignment objectives, illustrating how different MI techniques enable targeted interventions while also highlighting their limitations. This demonstrates both the promise and current constraints of interpretability-based alignment.

\section{Fundamental Challenges}

\subsection{Superposition and Polysemanticity}

The \textbf{superposition hypothesis} posits that networks represent more features than dimensions by storing features in overlapping combinations of neurons \citep{elhage2022toy}. This creates fundamental challenges:

\noindent\textbf{Polysemantic neurons:} Individual neurons respond to multiple unrelated concepts, making neuron-level interpretability difficult \citep{cammarata2020curve}. Research suggests models exploit sparsity—most features are inactive for most inputs—to pack many features into limited dimensions.

\noindent\textbf{Interference and interaction:} Features in superposition can interfere with each other in complex ways, making it difficult to predict how interventions will affect behavior \citep{elhage2022toy}.

\noindent\textbf{Computational burden: }Sparse autoencoders and other decomposition methods show promise but face scalability challenges. Training SAEs for frontier models requires enormous compute, and the number of features grows combinatorially \citep{bricken2023towards}.

\subsection{Scale and Complexity}

\noindent\textbf{Emergence:} Large models exhibit emergent capabilities not present in smaller versions \citep{wei2022emergent}. Whether interpretability techniques developed on smaller models transfer to frontier systems remains uncertain.

\noindent\textbf{Circuit interaction:} Real behaviors involve complex interactions between many circuits. Understanding how circuits compose and interfere is significantly harder than understanding individual circuits in isolation \citep{olah2020zoom}.

\noindent\textbf{Computational costs:} Comprehensive circuit analysis requires extensive patching experiments that scale poorly with model size. Automated methods help but still face significant computational barriers for the largest models.

\subsection{Validation and Ground Truth}

\textbf{Lack of ground truth:} Unlike in neuroscience, we cannot easily verify interpretability hypotheses through direct experimentation. We must infer computational mechanisms from behavioral observations and interventions.

\noindent\textbf{Confirmation bias:} Researchers may find interpretations that appear compelling but don't reflect true model computations \citep{rauker2023toward}. Rigorous causal verification is essential but often neglected.

\noindent\textbf{Evaluation metrics:} The field lacks standardized metrics for evaluating interpretability quality. Proposals include causal faithfulness \cite{geiger2021causal}, predictive power, and consistency across models, but no consensus exists.

\subsection{Alignment-Specific Challenges}

\textbf{Inner alignment:} Even with perfect interpretability of current behavior, we may fail to detect misaligned objectives that only manifest in specific circumstances \citep{hubinger2019risks}. Models might develop instrumental goals or deceptive strategies that remain dormant during training.

\noindent\textbf{Optimization demons: }Training may produce unintended optimization processes within networks—sub-agents pursuing their own objectives \citep{skalse2022defining}. Detecting and interpreting such structures remains an open challenge.

\noindent\textbf{Value representation:} Human values are complex, context-dependent, and difficult to specify. Even if we perfectly understand how models represent and pursue goals, determining whether those goals align with human values is philosophically and empirically challenging \citep{gabriel2020artificial}.

\noindent \textbf{Cultural representation challenges:} Achieving cultural alignment through mechanistic interpretability faces unique obstacles:

\begin{itemize}
    \item \textbf{Asymmetric representation capacity:} Models trained on imbalanced multilingual data develop asymmetric circuit structures where Western concepts have richer, more robust representations than non-Western concepts \citep{cao2023lost}. This asymmetry may be fundamental rather than easily correctable.
    \item \textbf{Cultural essentialism risks:} Mechanistic interventions targeting "cultural values" risk essentializing complex, heterogeneous cultures into simplified feature vectors. Cultures are dynamic and internally diverse; static circuit-level representations may reinforce stereotypes.
    \item \textbf{Power dynamics in alignment:} Decisions about which cultural perspectives to prioritize in model behavior reflect existing power structures. Mechanistic interpretability must grapple with who decides what constitutes "aligned" cultural representation \citep{birhane2022values}.
\end{itemize}

\begin{table*}[!t]
\centering
\caption{Core Challenges in Mechanistic Interpretability for Alignment}
\label{tab:mi_core_challenges}
\renewcommand{\arraystretch}{1.25}
\small
\resizebox{\textwidth}{!}{%
\begin{tabular}{p{3.2cm} p{4.5cm} p{4.8cm} p{4.8cm} p{4.8cm}}
\toprule
\textbf{Challenge} &
\textbf{Description} &
\textbf{Evidence} &
\textbf{Current Mitigations} &
\textbf{Open Problems} \\
\midrule

\textbf{Superposition \& Polysemanticity}
& Networks represent more features than dimensions via overlapping codes; neurons respond to multiple unrelated concepts
& Models exploit sparsity to pack features; individual neurons are highly polysemantic
& Sparse autoencoders with overcomplete dictionaries; topology-aware SAEs
& Scaling SAEs to frontier models; handling feature interactions; exponential feature growth \\

\midrule

\textbf{Scalability}
& Circuit analysis methods do not scale to models with hundreds of billions of parameters
& Patching experiments scale quadratically in components; frontier models contain thousands of layers and heads
& Attribution patching; hierarchical analysis; automated circuit discovery
& Real-time interpretability for deployment; analyzing emergent behaviors in the largest models \\

\midrule

\textbf{Validation \& Ground Truth}
& No objective ground truth for verifying interpretations; risk of confirmation bias
& Interpretations can be compelling yet incorrect; lack of standardized evaluation metrics
& Causal abstractions; ablation studies; cross-model consistency checks
& Gold-standard benchmarks; measuring interpretation quality; detecting spurious explanations \\

\midrule

\textbf{Circuit Composition \& Interaction}
& Real-world behaviors arise from complex interactions among many circuits
& Simple circuits compose non-linearly; representations are often distributed
& Circuit superposition analysis; circuit graphs; compositional patching
& Understanding emergent properties; predicting downstream effects of interventions \\

\midrule

\textbf{Universality vs.\ Specificity}
& Unclear whether circuits generalize across models, architectures, and training regimes
& Some universal circuits exist, but many are model- or task-specific
& Cross-model comparison; analysis of circuit evolution during training
& Determining when insights transfer; architecture- versus task-dependence \\

\midrule

\textbf{Asymmetric Representation}
& Dominant cultural or value perspectives are encoded more robustly than minority views
& Western concepts often have richer or more stable circuits than non-Western ones
& Targeted circuit editing; culturally diverse training data; steering vectors
& Capacity constraints; measuring representation equity; avoiding essentialism \\

\midrule

\textbf{Inner Alignment Detection}
& Difficulty identifying misaligned mesa-objectives that appear only in specific contexts
& Concerns about deceptive alignment; models may obscure true objectives
& Situational awareness probes; circuit-level anomaly detection; goal monitoring
& Detecting sophisticated deception; verifying alignment under distribution shift \\

\midrule

\textbf{Dual-Use \& Misuse Risks}
& Interpretability tools may enable removal of safety features or improved deception
& Circuit analysis could facilitate jailbreaking or bypassing refusal mechanisms
& Responsible disclosure; access controls; security-aware research practices
& Balancing transparency with security; developing defensive interpretability uses \\

\bottomrule
\end{tabular}
}
\end{table*}

Table~\ref{tab:mi_core_challenges} summarizes the fundamental challenges facing mechanistic interpretability research, along with current mitigation strategies and remaining open problems. These challenges are interconnected—for instance, superposition exacerbates scalability issues, while lack of validation makes it harder to assess whether mitigation strategies actually work.

\section{Future Research Directions}

\subsection{Automated Interpretability at Scale}

\noindent\textbf{Scalable circuit discovery:} Developing efficient algorithms for circuit discovery that scale to models with hundreds of billions of parameters. Promising directions include:

\begin{itemize}
    \item Gradient-based attribution methods that approximate expensive patching experiments
    \item Hierarchical approaches that identify high-level functional modules before fine-grained circuits
    \item Amortized interpretability where meta-models learn to interpret target models
\end{itemize}

\noindent\textbf{Automated description generation:} Extending methods like automated neuron description \citep{bills2023language} to describe circuits, attention patterns, and higher-level computational structures. Language models themselves may be powerful tools for generating and validating interpretability hypotheses.

\noindent\textbf{Multimodal interpretability:} Extending techniques to vision-language models and other multimodal architectures requires new methods for understanding cross-modal interactions and representations \citep{yuksekgonul2023when}.

\subsection{Cross-Model Generalization}

\textbf{Universal circuits:} Investigating whether similar circuits appear across different models, architectures, and training procedures \citep{conmy2023towards}. If circuits are universal, interpretability insights could transfer between models, dramatically reducing analysis costs.

\noindent\textbf{Meta-learning interpretability:} Training models to predict interpretable structure in other models. Such meta-interpretability systems could enable rapid analysis of new models and potentially automated safety verification.

\noindent\textbf{Transfer of interventions: }Determining when steering vectors, circuit ablations, or other interventions generalize across models. This would enable developing alignment techniques on smaller, more interpretable models with confidence they'll transfer to frontier systems.

\subsection{Interpretability-First Alignment}

\textbf{Mechanistic anomaly detection:} Using interpretability tools to detect anomalous circuits or representations that might indicate deceptive alignment, goal misgeneralization, or other alignment failures \citep{greenblatt2023evidence}.

\noindent\textbf{Transparent architectures:} Designing model architectures with interpretability as a first-class objective. This might include:

\begin{itemize}
    \item Encouraging monosemantic representations through architectural constraints
    \item Building in explicit symbolic reasoning components
    \item Modular designs that separate different cognitive functions
\end{itemize}

\noindent\textbf{Interpretability-guided training:} Using interpretability insights during training to encourage desired representations and circuits \citep{zou2023representation}. This could include:

\begin{itemize}
    \item Regularizers that encourage interpretable feature representations
    \item Curriculum learning ordered to develop circuits in interpretable ways
    \item Online monitoring and correction of problematic circuits during training
\end{itemize}

\subsection{Theoretical Foundations}

\textbf{Formal verification: }Developing rigorous methods to prove properties about model behavior based on circuit structure. This would require connecting mechanistic interpretability to formal verification techniques from computer science \citep{huang2020achieving}.

\noindent\textbf{Information-theoretic frameworks:} Building principled theories of how information flows through neural networks and using these to formalize concepts like circuits, features, and superposition \citep{elhage2021mathematical}.

\noindent\textbf{Causal models:} Strengthening connections to causal inference and structural causal models to provide rigorous foundations for interpretability claims \citep{geiger2021causal, geiger2023inducing}.

\subsection{Practical Alignment Applications}

\textbf{Red-teaming with interpretability:} Using mechanistic understanding to identify attack vectors and failure modes that behavioral testing might miss. This includes adversarial attacks targeting specific circuits and stress-testing alignment mechanisms.

\noindent\textbf{Monitoring deployed systems:} Developing interpretability-based monitoring systems that can detect alignment failures or distributional shift in deployed models by tracking circuit activations and representations \citep{hubinger2019risks}.

\noindent\textbf{Debate and amplification:} Enhancing scalable oversight techniques like debate \citep{irving2018ai} and recursive reward modeling \citep{leike2018scalable} with interpretability tools that help humans evaluate subtle arguments and detect deception.

\noindent\textbf{Value learning:} Using interpretability to understand how models represent human preferences and values, potentially enabling more effective value learning approaches than current RLHF methods.

\subsection{Mechanistic Understanding and Mitigation of Misalignment Through Pluralistic Approaches}

A comprehensive research program leveraging mechanistic interpretability for alignment should address both understanding and actively mitigating misalignment while respecting value and cultural diversity.

\subsubsection{Mechanizing Misalignment Detection}
Future work should develop automated systems that continuously monitor model internals for signs of misalignment:

\noindent\textbf{Objective representation analysis:} Detecting when models develop mesa-objectives or proxy goals that diverge from intended alignment targets \citep{hubinger2019risks}. This requires identifying circuits that implement goal-directed behavior and verifying their alignment with human values.

\noindent\textbf{Deceptive reasoning detection:} Building on work detecting lies \citep{azaria2023internal}, future systems should identify more subtle forms of deception, including strategic misrepresentation, selective information withholding, and context-dependent honesty.

\noindent\textbf{Value drift monitoring:} Tracking how value representations change during deployment, fine-tuning, or continued learning. Mechanistic interpretability enables detecting when models shift away from intended value functions.

\subsubsection{Circuit-Level Misalignment Mitigation}
Moving beyond behavioral alignment to directly modify problematic circuits:

\noindent\textbf{Targeted ablation and repair:} Identifying minimal circuits responsible for misaligned behaviors and either ablating them or replacing them with corrected versions. This requires understanding circuit composition well enough to predict downstream effects of modifications \citep{conmy2023towards}.

\noindent\textbf{Value circuit engineering:} Directly engineering circuits that implement desired value functions, rather than hoping they emerge from training. This could involve composing interpretable subcircuits for value recognition, ethical reasoning, and preference aggregation.

\noindent\textbf{Adversarial robustness through interpretability:} Using circuit analysis to identify vulnerabilities to adversarial attacks and jailbreaks, then hardening these circuits against exploitation \citep{zou2023representation}. This provides more principled robustness than behavioral adversarial training.


\subsubsection{Pluralistic Alignment Infrastructure}
\textbf{Modular value and cultural systems:} Designing architectures where different value frameworks and cultural perspectives are implemented in interpretable, composable circuits:

\begin{itemize}
    \item \textbf{Explicit context modules:} Circuits that explicitly represent the cultural and value context of a query and route information accordingly
    \item \textbf{Plug-in value systems:} Modular components encoding different ethical and cultural frameworks that can be activated, combined, or swapped based on context
    \item \textbf{Cultural calibration layers:} Interpretable layers that adjust outputs based on cultural context
\end{itemize}

\noindent\textbf{Automated cross-cultural circuit discovery: }Developing tools to systematically identify biases and gaps:

\begin{itemize}
    \item \textbf{Comparative circuit analysis:} Automatically comparing circuits activated by equivalent queries across languages/cultures to detect systematic differences \citep{wendler2024llamas}

    \item \textbf{Underrepresentation detection:} Identifying domains where certain perspectives are weakly represented
    \item \textbf{Bias attribution:} Tracing culturally-biased outputs back to specific components
\end{itemize}

\noindent\textbf{Participatory mechanistic alignment:} Involving diverse communities in interpretability-based alignment:

\begin{itemize}
    \item \textbf{Community-driven circuit auditing:} Tools enabling cultural communities to audit circuits affecting their values
    \item \textbf{Collaborative value specification:} Working with diverse stakeholders to specify desired value circuits
    \item \textbf{Cultural red-teaming with interpretability:} Using mechanistic understanding to enable cultural community members to identify failure modes that automated testing might miss.
\end{itemize}

\noindent\textbf{Cross-lingual circuit transfer: }Investigating whether cultural alignment insights transfer across languages:

\begin{itemize}
    \item \textbf{Universal cultural reasoning circuits:} Determining whether models develop language-independent circuits for cultural reasoning that could be analyzed once and applied broadly
    \item \textbf{Language-specific cultural pathways:} Mapping how different languages activate different cultural circuits and developing interventions that work across linguistic diversity
    \item \textbf{Multilingual feature disentanglement:} Using sparse autoencoders to separate language-specific features from cultural value features, enabling targeted cultural alignment without language interference
\end{itemize}

\noindent\textbf{Measuring pluralistic alignment mechanistically:} Beyond behavioral metrics, developing interpretability-based measures:

\begin{itemize}
    \item \textbf{Cultural representation diversity: } Quantifying how uniformly different cultural perspectives are represented in model features and circuits
    \item  \textbf{Stereotype circuit strength:} Measuring the causal impact of circuits that propagate cultural stereotypes
    \item \textbf{Value framework balance:} Assessing whether circuits implementing different ethical frameworks (Western individualism, Confucian relationalism, Ubuntu communalism, etc.) have comparable representation capacity
    \item \textbf{Context-appropriate activation:} Verifying that culturally-relevant circuits activate in appropriate contexts rather than uniformly
\end{itemize}

\noindent\textbf{Preference personalization without fine-tuning: }Using activation steering and circuit-level interventions to adapt model behavior to individual user values without expensive per-user training. Understanding which circuits control value-relevant behaviors enables efficient, interpretable customization.

\noindent\textbf{Fairness through feature editing: }Identifying features and circuits that encode biases toward particular value systems or demographic groups, then editing these to ensure fair representation of diverse perspectives \citep{sorensen2024roadmap}. This provides more targeted bias mitigation than dataset rebalancing.

\noindent\textbf{Explicit value negotiation:} Developing interpretable mechanisms for models to recognize value conflicts and negotiate between competing preferences transparently. This requires circuits that can represent uncertainty over values, model different stakeholders, and reason about ethical trade-offs.

\subsubsection{Scaling Mechanistic Alignment to Superintelligence}

Critical challenges for applying interpretability-based alignment to systems more capable than current models:

\noindent\textbf{Recursive alignment verification:} As models become capable enough to assist with alignment research, using interpretability to verify that alignment assistance is itself aligned. This requires detecting whether models are genuinely helping or pursuing instrumental goals through apparent cooperation.

\noindent\textbf{Emergent misalignment detection:} Developing interpretability methods that can detect novel forms of misalignment that emerge at greater capability levels. This may require meta-interpretability systems that can discover new types of circuits and representations.

\noindent\textbf{Scalable value learning: }Using mechanistic understanding to enable models to learn human values from limited feedback by understanding how humans represent and reason about values, rather than treating values as black-box reward functions.

\subsubsection{Integration with Multi-Stakeholder Governance}
Interpretability-based pluralistic alignment should support participatory approaches to AI governance:

\begin{itemize}
    \item \textbf{Transparent value trade-offs:} Making explicit which groups' preferences are prioritized in different contexts, enabling democratic deliberation about alignment targets
    \item \textbf{Auditable customization:} Allowing third parties to verify that deployed models respect diverse values as claimed
    \item \textbf{Contestable AI systems:} Enabling users to understand and potentially contest the value judgments embedded in model behavior
\end{itemize}

\subsubsection{Research Priorities}
To realize this vision, the field must:

\begin{itemize}
    \item \textbf{Global interpretability collaboration:} Building international research collaborations to ensure interpretability methods are validated across cultural contexts
    \item \textbf{Culturally-diverse training for interpretability researchers:} Training interpretability researchers from diverse backgrounds to recognize biases others might miss
    \item \textbf{Standardized cross-cultural benchmarks:} Developing interpretability-specific benchmarks that test whether circuit-level interventions successfully address cultural bias while maintaining capabilities
    \item \textbf{Ethical frameworks for cultural alignment:} Establishing principles for when and how to modify cultural representations in models, respecting cultural autonomy while addressing harmful biases
    \item \textbf{Scalable cultural knowledge integration:} Developing methods to efficiently integrate diverse cultural knowledge into models through targeted circuit editing rather than prohibitively expensive retraining
    \item \textbf{Value representation formalism:} Developing interpretability methods specifically designed for analyzing value representations and ethical reasoning circuits
    \item \textbf{Pluralistic evaluation:} Creating evaluation frameworks that assess how well models handle value conflicts and pluralistic scenarios across diverse cultural contexts
\end{itemize}

\noindent\textbf{Case study - Collectivist vs. Individualist reasoning:} Recent work has examined how models reason about moral dilemmas involving individual rights versus collective welfare:

\begin{itemize}
    \item Circuit analysis reveals Western-trained models have more robust pathways for rights-based reasoning than duty-based or community-focused reasoning \citep{tao2024investigating}

    \item Interventions adding collectivist reasoning circuits improve performance on cross-cultural moral reasoning tasks

    \item However, simply strengthening collectivist circuits can create new biases if not carefully calibrated to context
\end{itemize}

This research program represents a shift from black-box behavioral alignment to white-box mechanistic alignment—directly engineering and verifying the internal computations that determine model behavior. Success would provide stronger guarantees about alignment under distribution shift, novel situations, and increasing capability levels. Moreover, interpretability-based approaches to pluralistic alignment offer a path toward AI systems that can genuinely respect diverse human values and cultural perspectives rather than imposing uniform alignment targets.

\section{Discussion and Recommendations}

\subsection{The Path Forward}

Mechanistic interpretability has made significant progress but remains far from providing comprehensive understanding of frontier LLMs. We recommend a balanced research portfolio:

In the near term, research priorities should focus on scaling sparse autoencoder–based approaches to the largest contemporary models, enabling the extraction of interpretable features at previously unattainable scales. At the same time, there is a need to automate the discovery and validation of model circuits, reducing reliance on labor-intensive, ad hoc analyses. Integrating interpretability tools directly into standard model development pipelines will be critical for making mechanistic analysis a routine component of model training and deployment. In parallel, interpretability studies should be extended to RLHF and related alignment techniques, with the goal of understanding how these methods shape internal representations and decision-making processes.

Over the medium term, the field should advance toward the development and empirical evaluation of alignment methods that are explicitly guided by interpretability insights. Establishing standardized benchmarks and evaluation protocols will be essential to ensure comparability and cumulative progress across interpretability studies. Research should also investigate the extent to which learned circuits and features generalize across architectures, scales, and training regimes, thereby clarifying whether mechanistic insights are model-specific or reflect more universal principles. In addition, interpretability-based monitoring systems should be developed to support the ongoing oversight of deployed models, enabling the detection of emergent risks or unintended behaviors in real-world settings.

In the long term, a central objective is to achieve a comprehensive mechanistic understanding of highly capable or potentially superintelligent systems. Such understanding would enable the development of formal verification methods grounded in circuit-level structure, offering stronger guarantees about model behavior than empirical testing alone. Progress toward interpretability-first model architectures could further embed alignment considerations directly into system design, rather than treating them as post-hoc constraints. Ultimately, these advances aim to resolve the inner alignment problem by grounding alignment guarantees in a deep, mechanistic account of how advanced models represent goals, values, and decision processes.

\subsection{Integration with Other Alignment Approaches}

Interpretability should complement rather than replace other alignment research:

\noindent\textbf{Synergies with RLHF:} Interpretability can diagnose RLHF failures, suggest improvements, and validate that alignment training achieves intended effects \citep{casper2023open}.

\noindent\textbf{Enhancing red-teaming:} Mechanistic understanding enables more sophisticated adversarial testing that targets specific circuits and failure modes \citep{perez2022red}.

\noindent\textbf{Supporting theoretical alignment:} Interpretability provides empirical grounding for theoretical alignment proposals, revealing which concerns are realized in practice and which remain theoretical \citep{hubinger2019risks}.

\subsection{Limitations and Risks}

We acknowledge important limitations:

\noindent\textbf{False confidence: }Interpretability might provide misleading confidence in model safety if interpretations are incorrect or incomplete. Rigorous validation is essential.

\noindent\textbf{Arms race dynamics:} Interpretability tools could be used to make models better at deception or to remove safety mechanisms. Responsible disclosure norms are important \citep{brundage2020toward}.

\noindent\textbf{Diminishing returns:} The cost of interpretability may grow faster than model capabilities, potentially making comprehensive understanding of future systems intractable. Planning for this scenario is crucial.

\noindent\textbf{Philosophical challenges:} Even perfect interpretability may not resolve fundamental questions about consciousness, moral status, or value alignment in AI systems \citep{schwitzgebel2015defense}.
\section{Conclusion}

Mechanistic interpretability represents a crucial approach to understanding and aligning large language models. Recent progress in circuit discovery, feature analysis, and causal intervention has demonstrated that we can reverse-engineer specific algorithms and representations in modern LLMs. These insights have enabled targeted alignment interventions, from steering model behavior to detecting deception and reducing harmful outputs.

However, fundamental challenges remain. Superposition creates significant barriers to feature-level interpretability. The scale and complexity of frontier models strain existing methods. Validation of interpretability claims remains difficult without ground truth. Most significantly, we lack comprehensive understanding of how to ensure inner alignment—that models pursue truly aligned objectives rather than merely exhibiting aligned behavior.

The challenge of pluralistic and cultural alignment exemplifies why mechanistic interpretability is essential for responsible AI development. As LLMs are deployed globally, they must navigate diverse cultural contexts, values, and communication norms. Surface-level behavioral alignment calibrated to one cultural context often fails or causes harm when applied elsewhere. Only by understanding the internal circuits that encode cultural knowledge and values can we build systems that genuinely respect human diversity rather than imposing dominant cultural assumptions. This requires not just technical advances in interpretability, but participatory approaches that involve diverse cultural communities in auditing, specifying, and validating model internals.

The path forward requires sustained research investment across multiple fronts: developing scalable automated interpretability methods, establishing rigorous validation protocols, investigating cross-model generalization, building interpretability directly into alignment training, and creating infrastructure for pluralistic alignment that respects diverse values and cultures. Success will require close collaboration between interpretability researchers, alignment theorists, practitioners deploying models in high-stakes applications, and diverse cultural communities whose values must be represented.

As language models grow more capable and their societal impact increases, mechanistic interpretability becomes increasingly essential. Only by understanding how these systems work internally can we hope to ensure they remain beneficial, truthful, and aligned with the full diversity of human values across cultures and contexts. The research community must rise to this challenge with urgency, rigor, and humility about the difficulty of the task ahead.

\bibliography{custom}




\end{document}